\documentclass{article}
\usepackage{spconf,amsmath,graphicx}
\usepackage{bibspacing}
\usepackage{setspace}
\usepackage{javen}
\usepackage[pagebackref=true,breaklinks=true,letterpaper=true,colorlinks,bookmarks=false]{hyperref}

\title{AUTHOR GUIDELINES FOR ICASSP 2021 PROCEEDINGS MANUSCRIPTS}
\title{ISDA: Position-Aware Instance Segmentation with Deformable Attention}
\name{
Kaining Ying \qquad
Zhenhua Wang $^{\star}$  \qquad
Cong Bai  \qquad
Pengfei Zhou  \qquad
\thanks{$^{\star}$ indicates corresponding author (email: zhhwang@zjut.edu.cn).}
}
\address{College of Computer Science and Technology, Zhejiang University of Technology}
\begin{document}
%
\maketitle
\begin{abstract}
Most instance segmentation models are not end-to-end trainable due to either the incorporation of proposal estimation (RPN) as a pre-processing or non-maximum suppression (NMS) as a post-processing. Here we propose a novel end-to-end instance segmentation method termed ISDA. It reshapes the task into predicting a set of object masks, which are generated via traditional convolution operation with learned position-aware kernels and features of objects. Such kernels and features are learned by leveraging a deformable attention network with multi-scale representation. Thanks to the introduced set-prediction mechanism, the proposed method is NMS-free. Empirically, ISDA outperforms Mask R-CNN (the strong baseline) by 2.6 points on MS-COCO, and achieves leading performance compared with recent models. Code will be available soon.
\end{abstract}
\section{Introduction}
\label{sec:intro}

\begin{figure}[t!] 
	\centering 
	\includegraphics[width=0.45\textwidth]{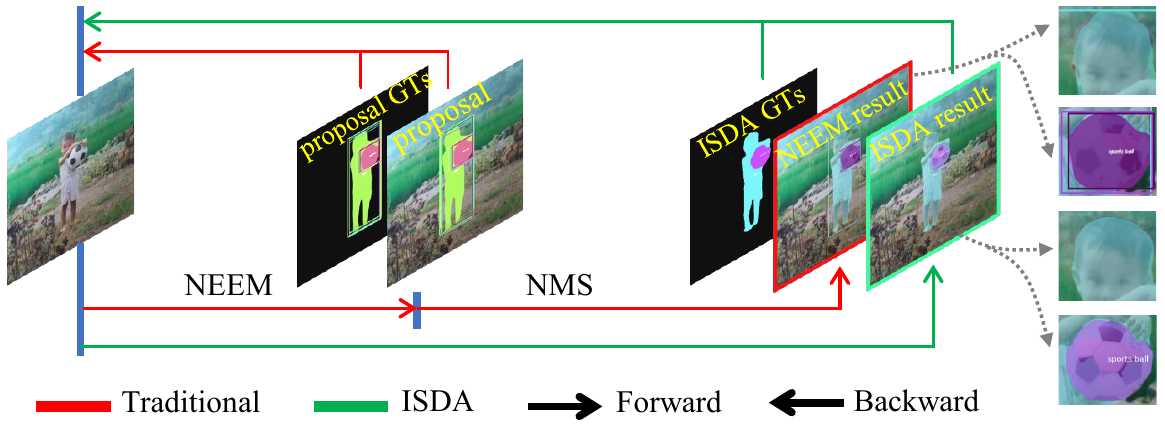} 
	\caption{Comparison of ISDA (ours) and traditional \textbf{N}on-\textbf{E}nd-to-\textbf{E}nd \textbf{M}odel (NEEM, \emph{e.g.,} Mask R-CNN) in training (left arrow) and testing (right arrow) phases, best viewed by zooming in. Note training and testing of NEEM  are not aligned due to usage of an extra NMS in testing. When redundant object proposals are provided, NEEM also gives redundant instance masks (see the football).  ISDA seldom produces redundant masks, and it typically gives clearly better results in border areas of objects (see the head of the child).} 
	\label{fig:compare}
\end{figure}

In vision community, removing hand-designed components and enabling the end-to-end training serve as stimulants to further performance improvement  \cite{ref:detr, ref:defodetr, ref:vistr, ref:end2end4, ref:end2end5, ref:end2end6}. 
In terms of instance segmentation \cite{ref:maskrcnn, ref:maskscorercnn, ref:FCIS, ref:panet, ref:htc, ref:solo, ref:solov2}, the main obstacles to train the model in an end-to-end way include two aspects. First, current segmentation methods, either top-down \cite{ref:maskrcnn, ref:FCIS, ref:panet, ref:maskscorercnn, ref:htc, ref:tensormask} or bottom-up \cite{ref:bottomup1, ref:bottomup2} have decomposed the task into several consecutive sub-tasks. Second, as shown by Fig.~\ref{fig:compare}, a post-processing step, namely NMS is typically taken to remove redundant predictions, which is non-differentiable and hinders back-propagating gradients. 

\begin{figure}[htbp] 
\centering 
\includegraphics[width=0.45\textwidth]{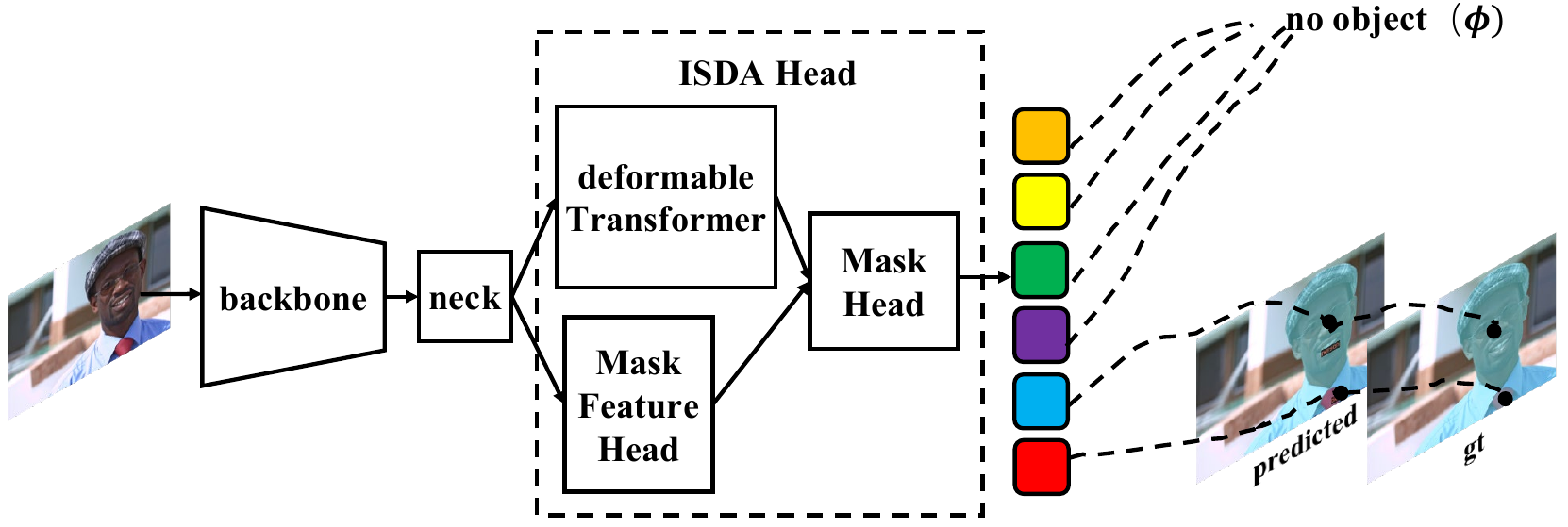} 
\caption{An overview of the proposed ISDA. The architecture consists of three ingredients: 
1) The backbone and neck module to extract multi-scale features;
2) The ISDA head which includes a deformable Tramsformer, a mask feature head and a mask head to predict object masks;
3) The Bipartite matching block which associates predictions with ground truth  to compute loss. All parameters of ISDA could be trained in an end-to-end fashion.}
\label{fig:overview} 
\end{figure}

Recently, DETR \cite{ref:detr} proposed to train an end-to-end detector with a set-based  loss and a Transformer encoder-decoder architecture. Nevertheless, DETR takes a long time to train due to the large computational overhead on dense attention computation, and it typically performs bad at detecting small objects as only a single-scale feature map is utilized. Very recently, deformable DETR \cite{ref:defodetr} introduced a sparse attention mechanism and used multi-scale features, which boosts the detection performance. Inspired by this, we craft an end-to-end instance segmentation framework termed ISDA, which is shown by Fig.~\ref{fig:overview}.

Similar to our proposed ISDA, SOLOv2 \cite{ref:solov2} also learns object kernels from data. ISDA has two merits over SOLOv2. First, instead of predicting object kernels for each cell in a fixed grid, ISDA is designed to adaptively learn object queries from data, which is more flexible than SOLOv2. Second, in order to overcome the translation-invariance of convolution, SOLOv2 introduces additional channels of relative coordinates for both kernel and feature learning. Apart from embedding positional information into feature learning, ISDA further concatenates the learned object positions (namely the reference points) with the object features to enhance the positional awareness of the learned object kernels. The effectiveness of such a design is assessed via ablation experiments.

Our main contributions are of three aspects. First, we propose an  instance segmentation framework based on deformable Transformer, which enables the end-to-end learning of object queries efficiently and effectively from data. Second, we present a method to generate position-aware object kernels which are especially useful for segmenting objects of similar appearances. Last but not least, our approach achieves leading performance on the challenging MS-COCO dataset.

\section{Related work}
\label{sec:related_work}

\subsection{Vision Transformer} Transformer \cite{ref:transformer} was first proposed for sequence-to-sequence machine translation and since then has become the {\itshape de~facto} standard in Natural Language Processing tasks. The core mechanism of Transformers is self-attention, which is beneficial in terms of modeling long-range relations. 
Transformers have shown to be promising in terms of addressing various vision tasks \cite{ref:detr, ref:defodetr, ref:fcobejctdetection, ref:sparsercnn, ref:vit, ref:setr, ref:sr, ref:videotr}. Very recently, VISTR \cite{ref:vistr} applied Transformer to instance segmentation in videos, a distinct task from our ISDA focusing on an image-level task. 

\subsection{End-to-End Instance-Level Recognition} An increasing number of works \cite{ref:detr, ref:defodetr, ref:fcobejctdetection, ref:sparsercnn, ref:ganbasedps, ref:cagnet} implement end-to-end detectors to achieve compelling results. To this end, the bipartite matching  \cite{ref:hungarian} has become the essential component for achieving end-to-end training of detectors. In the area of instance segmentation, this can be achieved by integrating sequential modules with recurrent networks \cite{REF:ISRA, ref:RIS}. Nevertheless, these early methods were only evaluated on small datasets without comparing against modern approaches. Recently, \cite{ref:istr} uses Transformer encoder only to fuse RoI features and image features to generate mask embedding. In contrast, ISDA uses deformable attention encoders and decoders to generate positional-aware kernels, which are then combined with mask features to generate masks directly.

\section{The ISDA Model}
\label{sec:ISDA}

As illustrated in Fig.~\ref{fig:overview}, ISDA contains three  blocks: a CNN backbone and neck, a ISDA head and a matching module to supervise the model training. This section first introduces the the backbone and neck of ISDA
\begin{figure}[htbp]
\centering
    \includegraphics[width=0.45\textwidth]{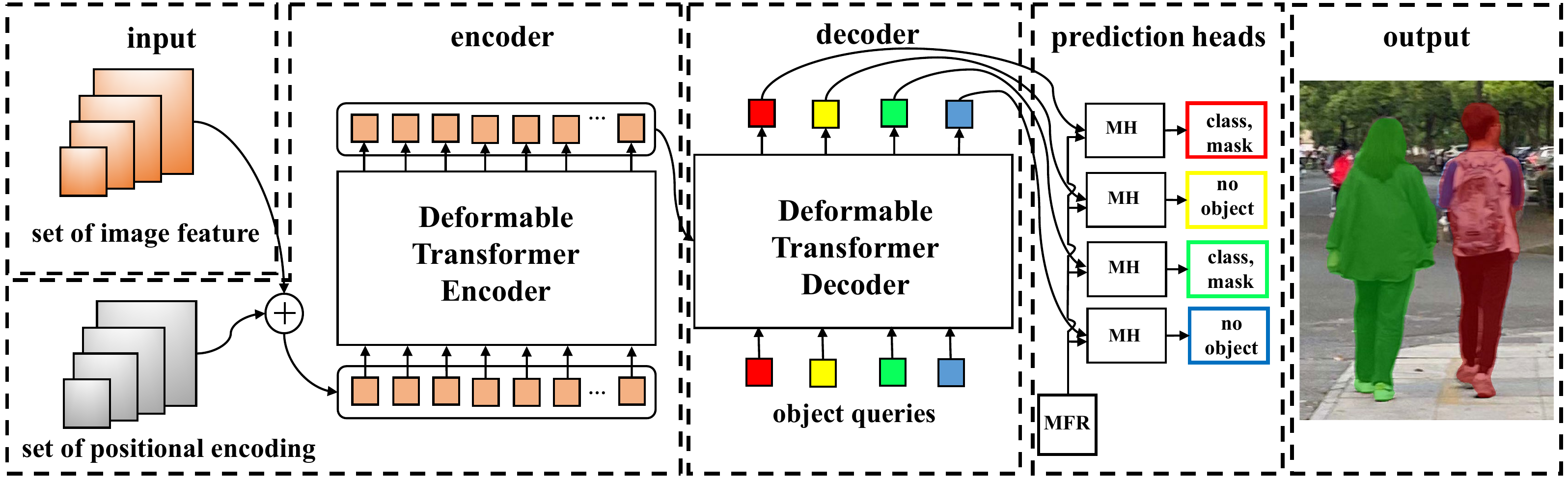}
    \caption{Illustration of the proposed ISDA head. ISDA takes as inputs the feature maps from neck supplemented with a set of encoded positions. We feed the inputs into deformable Transformer (encoder and decoder) to get a fixed amount of $(object-feature, ~reference-point)$ pairs. Then these pairs together with the mask feature representation (MFR) generated by a mask feature head (omitted here) are input into a series of mask heads (with shared weights) to generate final predictions. The generated masks are visualized in the rightmost diagram.}
    \label{fig:isda}
\end{figure}(Section~\ref{subsection3.1}). Section~\ref{sec:defTrans} elaborates the ISDA head, and Section~\ref{3.3} describes the bipartite match cost and the set prediction loss.

\subsection{Backbone and Neck}\label{subsection3.1}
Given an image denoted by $\xb \in \mathbb{R} ^ {3 \times H \times W}$, the CNN backbone extracts four feature maps with different resolutions, denoted by $\{C_i \in \mathbb{R}^{c_i \times H_i \times W_i}\}_{i=2}^{5}$. Here $c_i, H_i, W_i$ denote the channel number, the height and the width of the feature map $C_i$.
The neck takes the multi-scale features as the input and then enhances them separately as that done by deformable DETR \cite{ref:defodetr}. Consequently, we get $\{P_i \in \mathbb{R}^{256 \times H_i \times W_i}\}_{i=2}^{6}$, where $ P_6 $ is down-sampled form $C_5$.

\subsection{ISDA Head}\label{sec:defTrans}
Fig.~\ref{fig:isda} illustrates the architecture of our ISDA head. It contains  three components: an encoder-decoder deformable Transformer, a mask feature head (omitted in the figure) used to generate mask feature representations (MFR), and a mask head to make final predictions.

\textbf{Encoder.} We sum the feature maps $\{P_i\}_{i=3}^{6}$ and the encoded positions at different scales. This multi-scale representation is useful to identify which feature-scale the query pixel belongs to. Here queries and keys correspond to the embedded pixel-wise elements within the multi-scale feature maps. The outputs of the encoder take the same shapes as the inputs.

\begin{figure}[t!]
	\centering
	\includegraphics[width=0.48\textwidth]{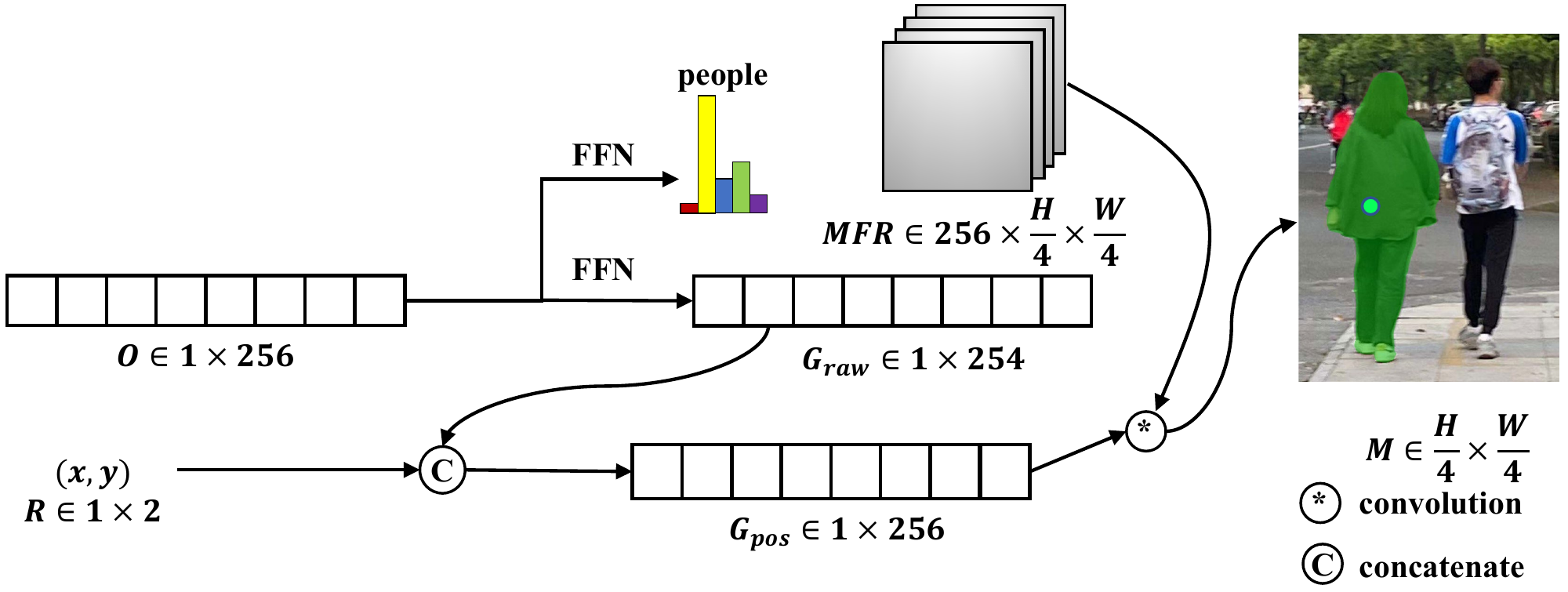}
	\caption{The structure of the mask head (best viewed by zooming in). The predicted mask (for left person only) and its reference point (the green circle) is visualized here.}
	\label{fig:MH}
\end{figure}

\textbf{Decoder.} The inputs of the decoder include the output from encoder and an extra object query vectors. Note that these queries are learned during training with random initialization, and are fixed for testing. Each layer includes two components, namely the cross-attention module and the self-attention module. The cross-attention module takes object queries to extract object features from input feature maps using a deformable attention manner. The self-attention module enables the object queries interacting with each other. The decoder outputs a set of object features and their corresponding reference points, which are taken to compute object kernels as depicted by Fig.~\ref{fig:MH}.

\textbf{Mask Feature Representation.} Inspired by SOLOv2, ISDA learns a compact and high-resolution mask feature representation (MFR) with feature pyramid. After repeated stages of $3 \times 3$ Conv, group-norm, ReLU and $2\times$ bilinear up-sampling, the neck features $\{P_i\}_{i=2}^{5}$ are fused (via element-wise summation) to create one single output at $1/4$ scale. It is worth noting that normalized pixel coordinates are fed into the smallest feature map (at $1/32$ scale) before convolution and upsampling. In ablation study in Section~\ref{sec:ablation}, we show the importance of appending such positional-information.

\begin{figure}[t]
	\centering
	\includegraphics[width=0.36\textwidth]{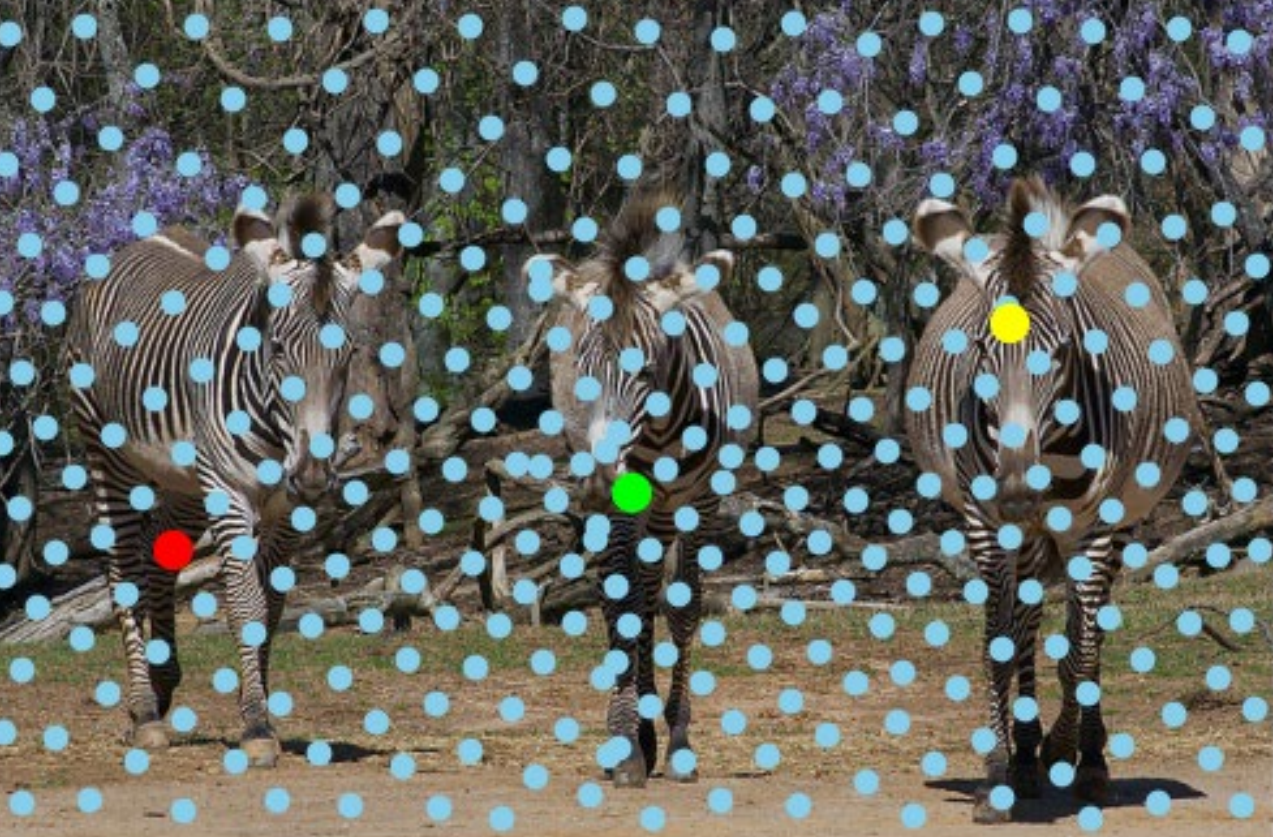}
	\caption{Visualization of learned reference points. Blue points represent background, while red, green and lemon points correspond to three zebras.}
	\label{fig:ref}
\end{figure}

\textbf{The Mask Head.} The Mask Head (MH) of ISDA is depicted by Fig.~\ref{fig:MH}. Its inputs include 1) the obejct feature vector $O \in \mathbb{R}^{256}$, 2) the normalized reference points $R$ (see an example in Fig.~\ref{fig:ref}) and 3) the mask feature representation $MFR \in \mathbb{R} ^ {256 \times H/4 \times W/4}$. Our MH generates all object masks in parallel in three steps. First, the object feature vector is fed into two different feed-forward networks (FFNs), in order to compute object classification scores $P_c$ and to obtain raw object kernel $G_{raw}$. Second, we concatenate $G_{raw}$ with $R$ to obtain the position-aware object kernel $G_{pos}$. Thanks to the the positional encoding in deformable Transformer, the object feature has included the positional information already. However, we find that appending the reference point to raw object kernel improves the position-awareness of the object kernel, hence is able to moderately improve the performance (see Section~\ref{sec:ablation}). Finally, $G_{pos}$ is convolved with $MFR$ to generate object masks $M$. 

\subsection{The Loss}
\label{3.3}
We follow the  procedure of \cite{ref:detr} in computing loss, which includes bipartite matching and loss computation. The only difference is that we replace the bounding box loss with the mask IoU loss, which is defined as $1 - {\rm IoU}(m_i, \hat{m}_{\sigma(i)})$. Here $m_i$ and $\hat{m}_{\sigma(i)}$ denote the $i^{th}$ ground truth and  predicted mask in a permutation $\sigma$. Please refer to \cite{ref:detr} for more details.

\section{Experiments}
\label{sec:experiment}

We implement ISDA based on the open source project MMDetection \cite{ref:mmdetection}. Unless otherwise specified, we use the ResNet-50 as our backbone network which are pretrained on ImageNet\cite{ref:imagenet}. The deformable Transformer and the classification branch in ISDA are pretrained on MS COCO for fast convergence. In the ablation experiments, all models are trained in 12 epochs with learning rate decay, which drops the learning rate by a factor at 9th and 11th epoch, respectively. For testing the final model is trained in 24 epochs with learning rate decayed at 16th and 22th epoch, respectively. Due to the limitation on GPU memories (only 3$\times$ GTX~1080~Tis are available), all compared models are trained by ourselves with the batch size equals 3. Following DETR\cite{ref:detr}, we use AdamW\cite{ref:adamw} optimizer and set the initial learning rate as 1.87e-5. We use a multi-scale training strategy and set  loss weights $\lambda_{cls}$ and $\lambda_{mask}$ to 1 and 3, respectively.


\begin{table}[t]
	\centering
	\caption{Results on MS-COCO \texttt{val2017} by changing the resolution of MFR. Here the ``resolution" column lists the ratio of the predicted mask to input image size. While increasing the resolution improves results on small objects ($AP_S$), it degrades results on large objects ($AP_L$).}\label{tab:resolution}
	\footnotesize
	\begin{tabular}{c|c|cc|ccc}
		\hline
		Resolution & $AP$   & $AP_{50}$ & $AP_{75}$ & $AP_S$  & $AP_M$  & $AP_L$  \\ \hline
		1/8        & 35.0          & 58.3 & 35.9 & 14.6 & 38.5 & \textbf{54.7} \\
		1/4        & \textbf{36.5} & \textbf{58.9} & \textbf{38.3} & 17.4 & \textbf{39.5} & 54.6 \\
		1/2        & 36.4          & 58.7 & 38.3 & \textbf{17.6} & 39.3 & 53.8 \\ \hline
	\end{tabular}
\end{table}

\subsection{Ablation Experiments}
\label{sec:ablation}

\begin{table}[t!]
	\centering
	\caption{Instance segmentation results with different auxiliary positional information. Including both MFR (MP) and Kernel positions (KP) delivers the best performance (with $\rm{Delta}=4.1$).}\label{tab:position}
	\footnotesize
	\begin{tabular}{c|c|c|c|cc|ccc}
		\hline
		MP & KP &Delta & $AP$    & $AP_{50}$  & $AP_{75}$  & $AP_S$   & $AP_M$   & $AP_L$  \\ \hline
		&                  & 0& 32.4  & 56.9  & 32.2  & 15.6  & 35.5  & 47.4 \\
		$\checkmark$   &         & +3.7& 36.1  & 58.5  & 37.9  & 16.6  & 39.0   & 54.5 \\
		& $\checkmark$     &-0.6& 31.8  & 56.0  & 31.9  & 15.4  & 34.8  & 47.1 \\
		$\checkmark$     & $\checkmark$&+4.1     & \textbf{36.5}  & \textbf{58.9}  & \textbf{38.3}  & \textbf{17.4}
		& \textbf{39.5}  & \textbf{54.6} \\ \hline
	\end{tabular}%
	\label{tab:addlabel}%
\end{table}%

To analyze ISDA, we conduct two ablation studies: 1) The choices of mask resolutions; 2) The positional information.

Recall that ISDA generates object masks by convolving the predicted object kernels with the mask features. Clearly, the 
mask resolution depends on the mask feature representation. We test three different resolutions which correspond respectively to $1/2$, $1/4$ and $1/8$ of the input image size, and the results are provided by Table~\ref{tab:resolution}. In general, $1/4$ scale admits the best performance (1.5 and 0.1 points better than $1/8$ and $1/2$ scales) in terms of average precision ($AP$). Not surprisingly, the highest resolution gives the best performance on small objects. However, it performs much worse on larger targets. Note all resolutions needs to be resized to the dimension of the original image, which inevitably results in the lost of details on object edges. As a trade-off, we use $1/4$ scale for all subsequent experiments because of its good performance and lower computational overhead. 

\begin{table}[t]
	\centering
	\caption{Comparison with Mask R-CNN (a strong baseline) and SOTA methods.}
	\label{tab:sota}
	\footnotesize
	\begin{tabular}{c|c|c|c|cc|ccc}
		\hline
		Method   & $AP$    & $AP_{50}$  & $AP_{75}$  & $AP_S$   & $AP_M$   & $AP_L$ \\ \hline
		Mask R-CNN \cite{ref:maskrcnn}           & 36.1  & 58.2  & 38.5  & 20.1  & 38.8  & 46.4 \\
		SOLO  \cite{ref:solo}           & 35.1  & 55.9  & 37.4  & 13.7  & 37.6  & 51.6 \\
		SOLOv2 \cite{ref:solov2}           & 37.4  & 58.4  & 40.1  & 15.4  & 40.2  & \textbf{57.4} \\
		CondInst \cite{ref:condinst}           & 36.9  & 58.2  & 39.6  & 19.8  & 39.3  & 48.0 \\
		BlendMask \cite{ref:blendmask}           & 37.0    & 58.0    & 39.4  & 19.5  & 39.9  & 53.1 \\
		ISTR \cite{ref:istr}                    & 37.6 & - & - & \textbf{22.1} & 40.4 & 50.6 \\
		ISDA (\textbf{ours})            & \textbf{38.7}  &\textbf{62.0}   & \textbf{41.1}  & 17.0  & \textbf{41.2}  & 55.7 \\ \hline
	\end{tabular}%
	\label{tab:addlabel}%
\end{table}

Aside positional encoding \cite{ref:transformer}, ISDA incorporates two extra sources of positional information to disentangle different objects taking similar appearance. The first source (MFR Pos.) is to add two channels of normalized pixel coordinates to the mask feature, as described in
Section~\ref{sec:defTrans}. The second source (kernel Pos.) is the reference points added into the object kernels, as illustrated by Fig.~\ref{fig:MH}. From Table~\ref{tab:position}, we can find that the baseline already has spatial awareness to some extent due to the zero-padding operation \cite{ref:solo}.
We can see that the performance improves by 3.7\% with MFR Positions. Note that kernel Pos. performs even worse than the baseline. This is because, without the aid of MFR Pos., the position-aware kernel (simply with kernel Pos.) is incapable of distinguishing similar objects due to the translation invariance of convolution. Interestingly, with the help of MFR Pos., kernel Pos. improves the result by 0.4\% (the bottom row) in terms of $AP$. Specifically, when doing convolution on MFRs, the concatenated coordinate channels in MFR and the reference points in kernels are multiplied to compute positional similarities. Hence the kernels are able to better identify the corresponding objects using both appearance and positional features. 

\begin{figure}[htbp]
\centering
		\includegraphics[width=0.47\textwidth]{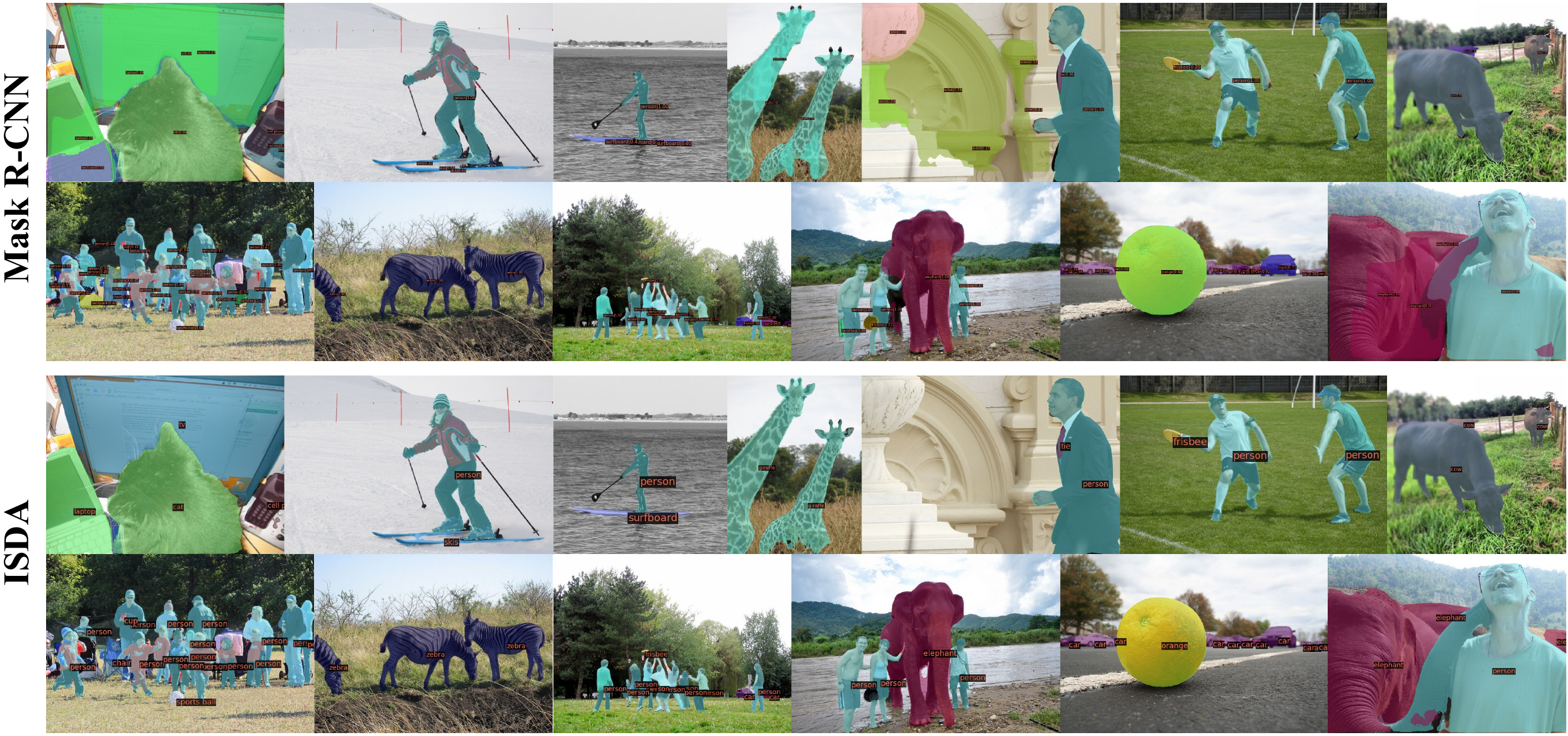}
	\caption{Mask R-CNN results \cite{ref:maskrcnn} (top) vs. ISDA results (bottom) on some examples. While Mask R-CNN gives duplicated masks and coarse edge segmentation results, our ISDA is able to address such issues and admits genuine masks. }
	\label{fig:vis}
\end{figure} 

\subsection{Qualitative and quantitative results}\label{sec:results}
We compare ISDA against SOTA mothods on MS-COCO \texttt{test-dev2017}, see Table~\ref{tab:sota} for details. Thanks to the introduced end-to-end training paradigm and the learned position-aware kernels, ISDA surpasses all SOTA approaches. Also note that ISDA outperforms Mask R-CNN \cite{ref:maskrcnn} (a well-known strong baseline) by 2.6 points in terms of $AP$, which is impressive in instance segmentation. While our ISDA performs best in general (on $AP$, $AP_{50}$ and $AP_{75}$), its results on small ($AP_S$) and large ($AP_L$) objects are not the bests. This is probably because ISDA uses deformable Transformer to sample and synthesize features across different scales, which benefits most the mask generation of middle-sized objects.

We visualize some segmentation results obtained by Mask R-CNN and ISDA in Fig.~\ref{fig:vis}. Mask R-CNN suffers from generating duplicated masks when NMS failed to filter out repeated instances, and it typically gives coarse masks along object edges. Thanks to the learned position-aware kernel and the informative mask features, ISDA is able to address such issues  and gives almost perfect segmentation, at least on these examples. However, ISDA can still get trouble in segmenting overlapping objects, as that shown by the rightmost diagram in bottom row. We will address this issue in  future. 

\section{Conclusion}
\label{sec:conclusion}
We have proposed a novel single-stage instance segmentation method, named ISDA. ISDA introduced a Transformer-style framework for instance segmentation, which effectively removed NMS and achieved end-to-end training and inference. Moreover, ISDA is able to distinguish similar objects better by learning  extra positional features. Empirically, ISDA admits genuine object masks and achieves leading performance compared with recent approaches.

\section*{Acknowledgement}
This work is supported by Zhejiang Provincial Natural Science
Foundation of China (LY21F020024, LR21F020002) and National Natural Science Foundation of China (61802348, U20A20196).

\begin{spacing}{1}
    \bibliographystyle{IEEEbib}
    \footnotesize
    \bibliography{strings,refs}
\end{spacing}

\end{document}